\documentclass[letterpaper, 10 pt, conference]{ieeeconf}  
\usepackage{float}
\usepackage{mathtools}
\usepackage{siunitx}
\usepackage{bbm}

\IEEEoverridecommandlockouts                              
\usepackage{cite}
\usepackage{array}
\usepackage{graphics} 
\usepackage{subfig}
\usepackage{multirow}
\usepackage{flushend}

\usepackage{url}
\usepackage{dsfont}
\usepackage{algorithmicx}
\usepackage[noend]{algpseudocode}
\usepackage{graphicx,caption}
\usepackage{amsmath,amssymb,stmaryrd,mathtools}
\usepackage{algorithm}
\usepackage{hyperref}
\usepackage{acronym}
\usepackage{color,xcolor}
\usepackage{verbatim}
\graphicspath{{./figures/}}

\newcommand{\ignore}[1]{}

\DeclareFontFamily{U}{MnSymbolC}{}
\DeclareSymbolFont{MnSyC}{U}{MnSymbolC}{m}{n}
\DeclareFontShape{U}{MnSymbolC}{m}{n}{
    <-6>  MnSymbolC5
   <6-7>  MnSymbolC6
   <7-8>  MnSymbolC7
   <8-9>  MnSymbolC8
   <9-10> MnSymbolC9
  <10-12> MnSymbolC10
  <12->   MnSymbolC12%
}{}
\DeclareMathSymbol{\powerset}{\mathord}{MnSyC}{180}

\definecolor{orange}{rgb}{1, .36, .08}
\definecolor{darkmagenta}{rgb}{0.698,0,0.698}

\definecolor{vg_edit_color}{rgb}{0, 0.0, 1.0}

\usepackage{todonotes}
\usepackage{soul}
\definecolor{smoothgreen}{rgb}{0.7,1,0.7}
\sethlcolor{smoothgreen}

\RequirePackage{luatex85}
\usepackage{pgfplots}
\pgfplotsset{compat=newest}
\pgfplotsset{every axis legend/.append style={%
cells={anchor=west}}
}
\usepgfplotslibrary{polar}
\usetikzlibrary{arrows}
\tikzset{>=stealth'}

\definecolor{C1}{rgb}{0.0, 0.447, 0.741}
\definecolor{C1_light}{rgb}{0.0, 0.6032388663967612, 1.0}
\definecolor{C2}{rgb}{0.85, 0.325, 0.098}
\definecolor{C3}{rgb}{0.929, 0.694, 0.125}
\definecolor{C4}{rgb}{0.494, 0.184, 0.556}
\definecolor{C5}{rgb}{0.466, 0.674, 0.188}
\definecolor{C6}{rgb}{0.301, 0.745, 0.933}
\definecolor{C7}{rgb}{0.635, 0.078, 0.184}

\usepgfplotslibrary{groupplots}

\usetikzlibrary{shapes.geometric, arrows}

\tikzstyle{startstop} = [rectangle, rounded corners, minimum width=2cm, minimum height=1cm,text centered, draw=black, fill=none]
\tikzstyle{arrow} = [thick,->,>=stealth]

\title{Improved Robustness and Safety for Autonomous Vehicle Control \\ with Adversarial Reinforcement Learning}

\author{Xiaobai Ma, Katherine Driggs-Campbell, and Mykel J. Kochenderfer
\thanks{
This material is based upon work supported by SAIC Innovation Center, a subsidiary of SAIC Motors.}%
\thanks{X. Ma, K. Driggs-Campbell, and M.J. Kochenderfer are with the Aeronautics and Astronautics Department, Stanford University, Stanford, CA, USA (e-mail: \{maxiaoba,krdc,mykel\}@stanford.edu).}%
}

\usepackage{cleveref}
\begin{document}

\maketitle

\begin{abstract}
To improve efficiency and reduce failures in autonomous vehicles, research has focused on developing robust and safe learning methods that take into account disturbances in the environment.
Existing literature in robust reinforcement learning poses the learning problem as a two player game between the autonomous system and disturbances.
This paper examines two different algorithms to solve the game, Robust Adversarial Reinforcement Learning and Neural Fictitious Self Play, and compares performance on an autonomous driving scenario.
We extend the game formulation to a semi-competitive setting and demonstrate that the resulting adversary better captures meaningful disturbances that lead to better overall performance.  
The resulting robust policy exhibits improved driving efficiency while effectively reducing collision rates compared to baseline control policies produced by traditional reinforcement learning methods.
\end{abstract}


\section{Introduction}


Robust control has a long history of providing guaranteed stability despite disturbances and modeling errors \cite{bacsar2008h}.
Such methods aim to achieve robust performance, given bounded disturbances or errors.
In a learning setting, the controls community has also used reachability tools to synthesize robust controllers \cite{akametalu2014reachability}.
These safe learning approaches provide guarantees on safe interaction and efficiency, but tend to be overly conservative \cite{gillula2013reducing,krdc2016}.

The reinforcement learning community has also addressed robustness through adversarial learning \cite{lowd2005adversarial,laskov2010machine}.
Instead of optimizing for expected reward, the system is optimized to reduce risk (i.e., maximize the rewards associated with the worst case trajectories \cite{tamar2015optimizing}). Traditional reinforcement learning methods sample trajectories from the simulator's natural distribution, meaning that the probability of dangerous rollouts is very low. Thus, even with a large negative reward on failures, the expected reward could still be high due to averaging. The rareness of such trajectories makes it hard to train effective polices, which is undesirable for risk-sensitive tasks. 

One solution is to bias the rollouts towards the worst case trajectories (e.g., trajectories from the tails of the disturbance distribution), and incorporate these samples in the training process. 
Rajeswaran et al. propose to sample a collection of trajectories from a source distribution and use a subset of low percentile trajectories to update the policy \cite{rajeswaran2016epopt}. 
This method effectively biases the training towards effectively handling extreme disturbances, but is not sample efficient. 
Another approach is to formulate the training as a two-player game between the system we wish to control and the environment disturbance. This method is referred to as Robust Adversarial Reinforcement Learning (RARL)~\cite{pinto2017robust}.

There is a trade-off between expected efficiency and robustness for safe driving.
Intuitively, if the worst case scenario is assumed in the design process, then the resulting controller will be overly cautious \cite{gillula2013reducing}.
In this work, we examine both worst-case performance, as is typically assessed for robust methods, as well as average reward to quantify this trade-off and compare different learning methods. 

While these methods address some safety problems, reinforcement learning approaches require simulation and therefore often fail to generalize to the real world.  
This drawback is apparent in adversarial settings where the adversary may not capture real-world disturbances.

There has been a great deal of recent research focusing on how to model and train the adversary from a learning and  game theoretic perspective.
One method called Fictitious Self-Play (FSP) solves for the Nash Equilibrium in this two players game using a reservoir buffer to train the adversary \cite{heinrich2015fictitious}.
Not only does this approach address some of the problems of overly cautious behavior and unrealistic disturbances by ``averaging" the adversarial disturbances used in simulation, but also provides theoretical guarantees.

The RARL and FSP approaches do not necessarily capture real-world disturbances, as we will demonstrate in this paper.
Moreover, these methods do not necessarily improve safe transfer for model mismatch and disturbances~\cite{pan2010survey}.

In this work, we extend these adversarial learning methods to a semi-competitive game setting, where the adversary in the two-player game is incentivized to adjust the disturbance magnitude according to the current capability of the control policy.
We show that this not only improves safety, but also improves robustness under different environment models.
We compare this trade-off between expected efficiency and safety for different game formulations and algorithms.

Finally, we apply our method to a transfer learning setting, where the training and testing are executed in different environments or with different dynamical models.
To do this, we propose a framework for training the system against an adversary that accounts for modeling errors and mismatch between the simulation used in training and real-world dynamics.
These real-world dynamics are derived from a test vehicle, which differs significantly from the model used to train the controller in simulation.
In doing so, we not only improve the performance, but also the computational efficiency of the learning and training process.
In summary, this paper presents the following contributions:
\begin{enumerate}
    \item We implement robust learning methods and assess the efficiency and safety trade-off for autonomous driving;
    \item We augment and improve existing game theoretic frameworks by adding non-competitive incentives to 
    actively adapt the disturbance magnitude;
    and 
    \item We demonstrate how these robust methods can account for model errors to improve transfer learning from simulation to real-world dynamics.
\end{enumerate}


\section{Problem Formulation}
\label{sec:related_work}

Suppose we would like to control a vehicle governed by a dynamical equation:
\begin{equation}
\label{eq:vehicle dynamics}
x[k+1] = f(x[k],u[k])+g(x[k],u[k],d[k])
\end{equation}
where $f$ is the dynamics of the vehicle, $x[k]\in\mathbb{R}^n$ is the vehicle state at time $k\in\mathbb{N}$, $u[k]\in\mathbb{R}^m$ is the control input, and $g$ is a function that captures the uncertainty in the system and the external environment as a function of the state, input, and disturbances $d[k]$. 

A precise dynamical model may be unknown or difficult to compute. We approximate the vehicle dynamics with a simple discrete time bicycle model and noise model:
\begin{equation}
\label{eq:simulation model}
x[k+1]=\hat{f}(x[k],u[k])+v(x[k],u[k],d[k])    
\end{equation}
where $\hat{f}$ is an approximate model of the true dynamics, $v$ is a function that attempts to capture modeling error and environment disturbances, and all other variables are as previously described. \Cref{sec:exp_setup} provides further details.

One approach to train a policy that is robust to model mismatch and noise is by introducing a two-player Markov game \cite{pinto2017robust}, which is a special case of a stochastic, dynamic game, for which one or more players have probabilistic transitions between states.  In the Markov game, it is assumed that the distribution of next state of the game only depends on the current state and action.
In this formulation, the modeling error and uncertainties are modeled as external disturbances. Player 1 (the protagonist) is trained to effectively control the vehicle, and the external disturbances are generated by player 2 (the adversary).


\subsection{Game Theoretic Formulation}

This two player game can be expressed as a tuple $(S,A_1,A_2,P,r_1,r_2,\gamma,s_0)$, where $S$ is the state space of the vehicle initialized at $s_0$; $A_1$ is the action space for protagonist, which will be the acceleration and steering commands ($u[k]$ in \cref{eq:simulation model}); $A_2$ is the action space for the adversary, which will be the disturbances 
($v(\cdot)$ in \cref{eq:simulation model}); $P:S \times A_1 \times A_2 \times S \rightarrow \mathbb{R}$ is the transition probability density defined by equation \cref{eq:simulation model}; $r_1: S \times A_1 \times A_2 \rightarrow \mathbb{R}$ and $r_2: S \times A_1 \times A_2 \rightarrow \mathbb{R}$ are the rewards for protagonist and adversary; and $\gamma$ is a discount factor.

The reward function determines the type of game and has a significant impact on equilibria, and therefore performance of the resulting policy. We consider the following two types of games: strictly competitive and semi-competitive.

\subsubsection{Strictly Competitive Games} In the existing literature on robust adversarial learning, the game is generally posed as a zero-sum or strictly competitive game, where the protagonist and the adversary are in direct opposition \cite{pinto2017robust,heinrich2015fictitious}.  
Specifically, the reward for the adversary is the opposite of the protagonist: $r_1 = -r_2$, where the protagonist is maximizing $\gamma$-discounted rewards while the adversary is minimizing them. 
With this setup, the adversary is helping to sample worst case trajectories for the protagonist. 
In RARL, the adversary's role is to sample trajectories from the worst $\alpha$-quantile.
The $\alpha$ parameter is determined by the magnitude of the disturbance available to the adversary~\cite{pinto2017robust}.
One drawback of a zero-sum game is that a disturbance at maximum magnitude is almost always the best output for the adversary, as it disrupts the system most. 

In our experiments, as will be presented in \Cref{sec:robust vali,sec:realworld results}, we observed three key different failures caused by training in a strictly competitive setting: (1) The protagonist becomes too conservative, thus exhibiting poor performance.; (2) The protagonist is unable to recover from failures early in the training processing and falls into a local minimum at a low performance region; and (3) The adversary becomes predictable by consistently outputting maximal disturbance, causing the protagonist to overfit and thus poorly generalizes over different disturbance distributions. 

\subsubsection{Semi-Competitive Games}
We propose augmenting the traditional zero-sum game approach to a semi-competitive or nonzero-sum setting, where the adversary has both a competitive and a cooperative component in its reward function.
Formally, we change the adversary reward to $r_2 = -r_1+r_c$, where $r_c$ is a cooperative reward that actively regulates the disturbance magnitude.

In the early stages of training when the protagonist is relatively weak, disturbances with different magnitude receive similar competitive reward $-r_1$. To earn some cooperative rewards, the adversary would prefer weak disturbances. As the protagonist iteratively learns and becomes more robust, a weak disturbances receives less competitive reward, while a stronger disturbance with higher competitive reward but lower cooperative reward is preferable.  
Thus, the adversary would need to trade-off between the competitive and cooperative rewards.  
The desired disturbance distribution would aim to cause a failure while using the minimal disturbance magnitude. 
This disturbance is also desired for the protagonist since it is easier for the protagonist to recover. 
By adding a cooperative reward, we make the adversary adaptively adjust its disturbance magnitude such that it allows the protagonist to effectively learn, while being adversarial.

While there are many forms which $r_c$ may take, for the adversary that outputs disturbances following a Gaussian distribution, we found that a simple rectangle function $r_c=r_a \cdot \mathds\{|d[k]|<d_{max}\}$ works well, where $d[k]$ is the disturbance to be applied, $d_{max}$ is the maximum allowed disturbance, and $r_a$ is  a positive constant. 
The magnitude of $r_a$ is a weighting between the competitive and cooperative reward, and is inverse related to the $\alpha$-percentile that would be sampled. 

\subsection{Solution Methods}
To solve this high-dimensional, complex game and train a policy with reinforcement learning, we implement two methods to find the equilibria in the game: Robust Adversarial Reinforcement Learning (RARL) and Neural FSP (NFSP).

\subsubsection{RARL}  
As originally proposed by Pinto et al., this approach uses neural network policies to represent each of the players and iteratively trains the protagonist and the adversary against each other, alternating at each training epoch \cite{pinto2017robust}.
The adversarial network learns to perturb the vehicle trajectories to maximize its reward. 
Then, by optimizing the protagonist against this evolving adversary, hypothetically, the protagonist learns to perform robustly over the entire disturbance space. 
The training algorithm is shown in \cref{Alg:RARL} assuming the protagonist and adversary follow stochastic policies $\pi_{P}$ and $\pi_{A}$. 

\begin{algorithm}[t!]
\caption{RARL Training Procedure \cite{pinto2017robust}}
\label{Alg:RARL}
\begin{algorithmic}
\State \textbf{Input:} Environment $E$; Stochastic policies $\pi_{P}^{(0)}$ and $\pi_{A}^{(0)}$
\For {$i=1,\dots,N_{iter}$}
\State $\pi_{P}^{(i,0)}\leftarrow\pi_{P}^{(i-1)}$
\For {$j=1,\dots,N_1$}
\State $paths \leftarrow \Call{Rollout}{E,\pi_{P}^{(i,j-1)},\pi_{A}^{(i-1)}}$
\State $\pi_{P}^{(i,j)} \leftarrow \Call{PolicyOptimizer}{\pi_{P}^{(i,j-1)},paths}$
\EndFor
\State $\pi_{P}^{(i)}\leftarrow\pi_{P}^{(i,N_1)}$
\State $\pi_{A}^{(i,0)}\leftarrow\pi_{A}^{(i-1)}$
\For {$k=1,\dots,N_2$}
\State $paths \leftarrow \Call{Rollout}{E,\pi_{P}^{(i)},\pi_{A}^{(i,j-1)}}$
\State $\pi_{A}^{(i,j)} \leftarrow \Call{PolicyOptimizer}{\pi_{A}^{(i,j-1)},paths}$
\EndFor
\State $\pi_{A}^{(i)}\leftarrow\pi_{A}^{(i,N_2)}$
\EndFor
\end{algorithmic}
\end{algorithm}

\subsubsection{NFSP}
Fictitious Self-Play is a method for finding the Nash equilibrium in two-player, multi-step games \cite{heinrich2015fictitious}. 
Instead of iteratively training against an optimized opponent, players choose best responses to other players average behavior. 
Neural Fictitious Self-Play (NFSP), listed in \cref{Alg:NFSP}, combines FSP with neural network polices to minimize divergence in common multi-agent reinforcement learning methods \cite{heinrich2016deep}. 
An agent consists of two neural networks: one network to learn an approximate best response to the average response of other agents, and a second network that averages the agent's own historical strategies.
The average response is determined by using a reservoir buffer to store the agent's past behavior.


\begin{algorithm}[t!]
\caption{NFSP Training Procedure}
\label{Alg:NFSP}
\begin{algorithmic}
\State \textbf{Input:} Environment $E$
\State \textbf{Initialize:} Reservoir buffers $M_{P}$ and $M_{A}$; best response policies: $\pi_{P,br}^{(0)}$, $\pi_{A,br}^{(0)}$; average policies: $\pi_{P,avg}^{(0)}$, $\pi_{A,avg}^{(0)}$
\For {$i=1,\dots,N_{iter}$}
    \State $\pi_{P,br}^{(i,0)}\leftarrow\pi_{P,br}^{(i-1)}$
    \For {$j=1,\dots,N_1$}
    \State $paths \leftarrow \Call{Rollout}{E,\pi_{P,br}^{(i,j-1)},\pi_{A,avg}^{(i-1)}}$
    \State $\pi_{P,br}^{(i,j)} \leftarrow \Call{PolicyOptimizer}{\pi_{P,br}^{(i,j-1)},paths}$
    \State $\Call{Populate}{M_{P},paths}$
    \EndFor
    \State $\pi_{P,br}^{(i)}\leftarrow\pi_{P,br}^{(i,N_1)}$
    \State $\pi_{P,avg}^{(i)} \leftarrow \Call{Fit}{\pi_{P,avg}^{(i-1)},M_{P}}$
    \State $\pi_{A,br}^{(i,0)}\leftarrow\pi_{A,br}^{(i-1)}$
    \For {$k=1,\dots,N_2$}
    \State $paths \leftarrow \Call{Rollout}{E,\pi_{P,avg}^{(i)},\pi_{A,br}^{(i,j-1)}}$
    \State $\pi_{A,br}^{(i,j)} \leftarrow \Call{PolicyOptimizer}{\pi_{A,br}^{(i,j-1)},paths}$
    \State $\Call{Populate}{M_{A},paths}$
    \EndFor
    \State $\pi_{A,br}^{(i)}\leftarrow\pi_{A,br}^{(i,N_2)}$
    \State $\pi_{A,avg}^{(i)} \leftarrow \Call{Fit}{\pi_{A,avg}^{(i-1)},M_{A}}$
\EndFor
\end{algorithmic}
\end{algorithm}
\section{Experimental Methods}
\label{sec:exp_setup}

This section presents the experiments designed to train and test the different robust adversarial reinforcement learning algorithms as well as the implementation details.

\subsection{Experimental Setup}
\label{subsec: exp_setup}
\begin{figure}[t!]
\vspace{-5pt}
    \centering
    \includegraphics[width = .9\columnwidth]{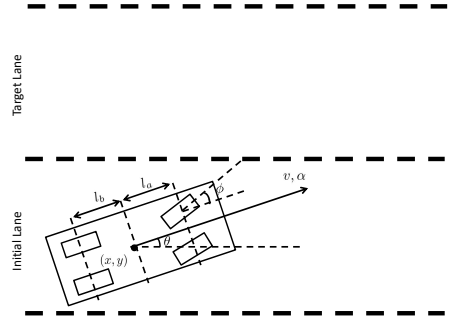}
    \caption{\small Scenario representation of lane changing experiment setup.}
    \label{fig:setup}
    \vspace{-5pt}
\end{figure}

We hypothesize that using robust learning techniques will not only lead to safer control policies, but will also improve performance when transferring this policy to the real-world system, which will likely operate with different dynamics. To study this, we created a training environment where the protagonist controls the ego vehicle for lane keeping or lane changing. 
The ego vehicle drives on a straight three-lane roadway with infinite length, and there are no other vehicles on the road. The simulation environment is built upon a 2D traffic simulator, AutomotiveDrivingModels.jl.\footnote{https://github.com/sisl/AutomotiveDrivingModels.jl} The vehicle is initialized in the center lane with an initial velocity of 10 m/s. 
At the beginning of each trial, a target lane is randomly assigned. 
The time step of the simulation environment is 0.05s, and the maximum time duration of each path is 10s. 
This experiment is illustrated in \cref{fig:setup}.

The Markov decision process used in the Markov game is defined as follows:

The state of the vehicle $s_t\in S$ is represented as $s_t=[x_t,y_t,v_t,\theta_t]$, where $(x_t, y_t)$ is the position, $v_t$ is the speed, and $\theta_t$ is the orientation of the vehicle.

The action of the protagonist $a_{1,t}$ is a two dimensional vector representing the acceleration and steering of the ego vehicle at each timestep, where $a_{1,t} = [\alpha_t,\phi_t]$. The acceleration is limited to \num{-3} to \SI{3}{\meter\per\square\second}, and the steering is limited to \ang{-20} to \ang{20}.

The action of the adversary $a_{2,t}$ is the disturbance on the acceleration and steering, $a_{2,t} = [\delta \alpha_t, \delta\phi_t]$. The magnitude of the disturbance is limited to 20\% of the maximum acceleration and steering.

The state transition model, as generally stated in \cref{eq:simulation model} and visualized in \cref{fig:setup}, follows a simple discrete time bicycle model, where $l_a=l_b=\SI{1.5}{\meter}$ are distances between the vehicle's center of mass and its front and rear axle, $\Delta t=\SI{0.05}{\second}$ is the time step, and all other variables are as previously described.

The reward function of the protagonist $r_{1,t}$ is a weighted sum of the rewards on vehicle state and action:
  \begin{equation}
  r_{1,t} = 0.5r^{v}_t+0.5r^{\theta}_t+0.1r^{\alpha}_t+0.2r^{\phi}_t+1.0r^{y}_t    
  \end{equation}
  where the reward on the velocity is given by a smooth, increasing function with range $[-1,1]$:
  \begin{equation*}
  r^{v}_t = \log_{\frac{v_{max}-v_{min}}{2}}(100 \cdot \frac{v_t-v_{min}}{v_{max}-v_{min}}+0.99)-1
  \end{equation*}
  The reward associated with orientation is given by:
  \begin{equation*}
      r^{\theta}_t=\begin{cases} 
      -1.0 & \text{ if } |\theta_t|>\pi/4\\
      0.0 & \text{ otherwise}
   \end{cases}
  \end{equation*}
  The reward associated with the acceleration is given by:
  \begin{equation*}
      r^{\alpha}_t=-\frac{|\alpha_t|}{\alpha_{max}}
  \end{equation*}
  The reward associated with the steering is given by:
  \begin{equation*}
      r^{\phi}_t=-\frac{|\phi_t|}{\phi_{max}}
  \end{equation*}
  The reward associated with the vehicle position is determined by the lateral offset: 
  \begin{equation*}
      r^y_t = \begin{cases} 
      3.0 & \text{ if } |y_t-y_{goal}|<0.05\\
      1.0-\frac{|y_t-y_{goal}|}{l_w} & \text{ otherwise }
   \end{cases}
  \end{equation*}
where $y_{goal}$ is the lateral position of the center line of the target lane and $l_w=\SI{3}{\meter}$ is the lane width. If the ego vehicle goes off the road, meaning that a failure and collision has occurred, $r_{1,t} = -5.0$ and the trajectory ends.

In the zero-sum setting, the adversary's reward is strictly competitive, meaning $r_{2,t}=-r_{1,t}$. In the nonzero-sum setting, the adversary's reward is given by $r_{2,t}=-r_{1,t}+r_{c,t}$, where $r_{c,t}$ is the additional adversary reward taking the form: 
\begin{subequations}
\begin{align}
  r_{c,t} =& r_{\delta\alpha,t}+r_{\delta\phi,t}\\
  r_{\delta\alpha,t} =&
      \begin{cases}
      r_a & \text{ if } |\delta\alpha_t| < \delta \alpha_{max} \\
      0 & \text{ otherwise } 
      \end{cases}\\
  r_{\delta\phi,t} =&
      \begin{cases}
      r_a & \text{ if } |\delta\phi_t| < \delta \phi_{max} \\
      0 & \text{ otherwise } 
      \end{cases}
\end{align}
\end{subequations}
  where the maximum disturbance limit, $\delta\alpha_{max}$ and $\delta\phi_{max}$,  is predefined. Note that $r_a=0$ means a zero-sum setting. In the experiment, $r_a$ is set to 3.\footnote{The magnitude of $r_a$ is inversely related to the $\alpha$-percentile that would be sampled.  We ran several experiments on different $r_a$ values and found that 3 gives the best result.}
\subsection{Implementation Details}
To compare the performance of different policies, we train five different networks.
First, we train a baseline policy using traditional reinforcement learning.  
Then, we train policies using the RARL and the NFSP methods in a strictly competitive and a semi-competitive settings.
We used the following parameters for training.

\subsubsection{Baseline Policy}
The baseline policy is trained with Trust Region Policy Optimization (TRPO) as implemented by RLLab with no disturbances in the environment \cite{schulman2015trust,duan2016benchmarking}. 
The baseline uses a Gaussian Multilayer Perceptron architecture with hidden layer sizes of 256, 128, 64, and 32 with ReLu nonlinearity activation functions.
The training is conducted over 15000 iterations with a batch size of 4000 and a step size of 0.01. 

\subsubsection{RARL}
For training the policies with RARL, we use the  same policy structure as the baseline for both the protagonist and the adversary. 
The protagonist is initialized with a baseline policy trained with 5000 iterations, and the adversary is randomly initialized. The training is conducted over 4000 iterations, and $N_1=N_2=5$, which is the number of policy updates performed for the protagonist and adversary policies in each iteration. The batch size is 400. 
The policy optimizer and other parameters are consistent with the baseline.
The training parameters are chosen so that the training time for the baseline (starting from 5000 iterations) and RARL is approximately the same.

\subsubsection{NFSP}
For NFSP training, we use the same network structure as the baseline for the protagonist and adversary's best response policy, $P_{br}$ and $A_{br}$. The average response policies, $P_{avg}$ and $A_{avg}$, use RLLab's Gaussian Multilayer Perceptron Regressor implementation. The hidden layer sizes and nonlinearities are consistent with the baseline. 
The reservoir buffers $M_P$ and $M_A$ have size \num{2e4}. Other parameters are consistent with RARL.

\subsection{Validation Dynamics}
\label{subsec: validation dynamics}
\begin{figure}[!t]
\vspace{7pt}
    \centering
    \includegraphics[width = .85\columnwidth]{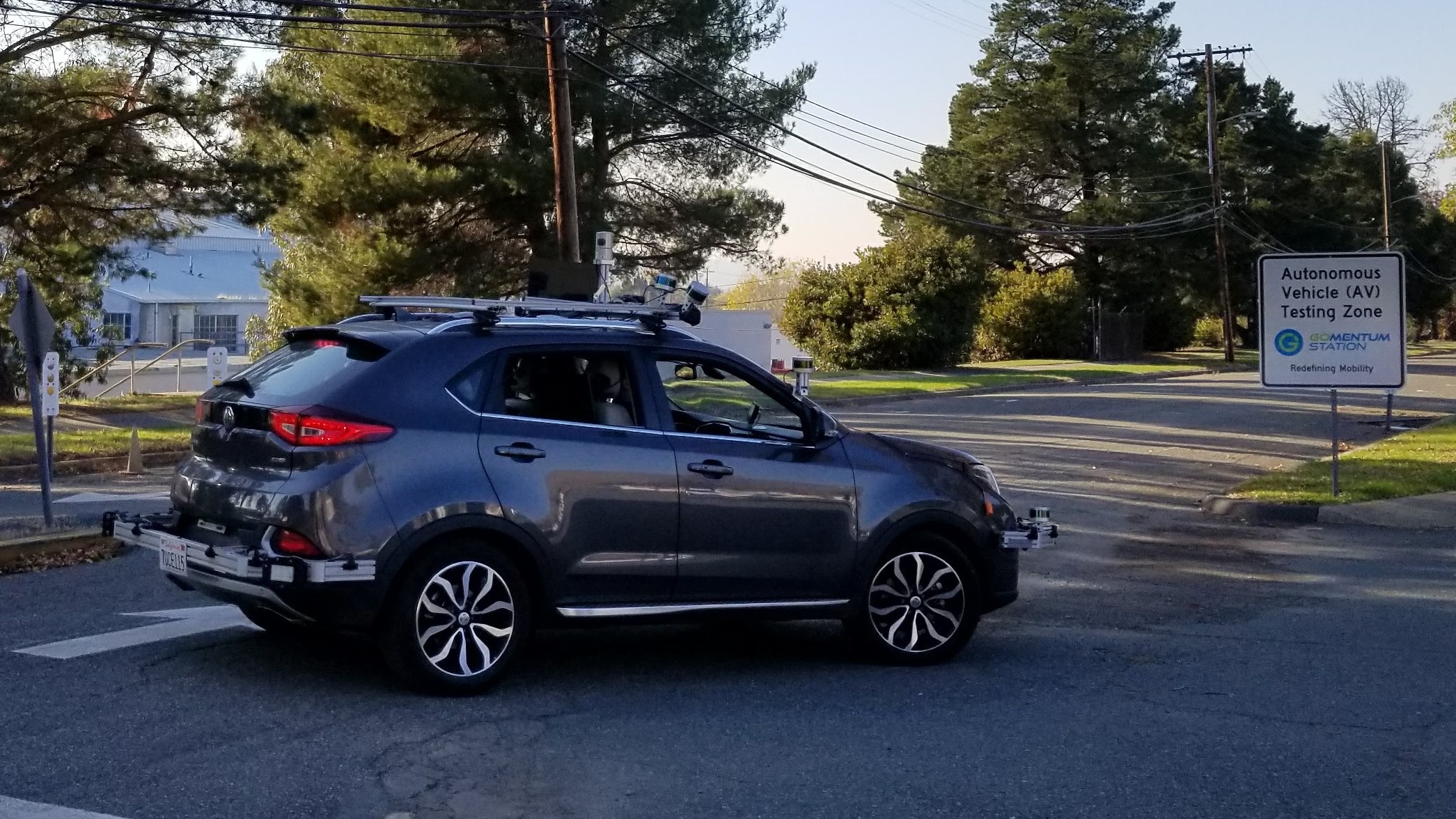}
    \caption{\small Test vehicle provided by SAIC Innovation Center.}
    \label{fig:vehicle}
\vspace{-17pt}
\end{figure}

We hypothesize that using the proposed robust methods will improve performance when the policy is executed on a real vehicle, despite modeling errors and noise.
This hypothesis is validated using a dynamical model closer to the test vehicle provided by the SAIC Innovation Center. \Cref{fig:vehicle} shows the test vehicle.

While testing on the vehicle, we observed that much of the model mismatch was caused by the steering response of the vehicle. In training, we assume that the steering of the vehicle perfectly tracks the the command steering, which is not true in practice. 
The limits on the steering and angular rate is an important constraint that is often tuned and modified in the low-level controller on the vehicle.  

For vehicular control, there are many vehicle specific model parameters that significantly impact the dynamics, such as the axle distance ($l_a$ and $l_b$ shown in \cref{fig:setup}). 
The resulting policy should be robust to model mismatch with respect to changes in these parameters so that it is applicable on different vehicles.

\subsection{Evaluation Metric}
\label{subsec:metric}
To test the control policy, we run each policy for 500 rollouts.  We need two metrics to evaluate the efficiency and the safety of the polices. For efficiency, we use the the average undiscounted path reward introduced in \cref{subsec: exp_setup}, which is a combined metric on difference aspects of the driving. We subtract the collision part of the reward to make it orthogonal to the safety metric. For safety, we use the collision rate which is the number of failures divided by the number of rollouts.

\section{Robustness to Adversarial Disturbances}
\label{sec:robust vali}
To validate our methods and demonstrate robustness, we conduct several tests with different disturbance distributions and vehicle dynamics to evaluate the performance of the trained protagonists.
First, we test the robust policy with a disturbance that follows a Pareto distribution.
Second, we show robustness to an adversarial disturbance. 

\subsection{Pareto Disturbance Test}
To validate the systems' robustness, we test the policy using a heavy tailed distribution.  
The disturbances on acceleration and steering are sampled from a Pareto distribution with shape parameter $\beta$ that varies from 1 to 10 and scale parameter $x_m=1$.\footnote{This distribution gives the normalized disturbance magnitude.  In testing, this value is scaled to 20\% of the maximum acceleration or steering.  Additionally, half of the tests use the opposite of the sampled disturbance.} 
The probability density function is $\frac{\beta x_m^\beta}{x^{\beta+1}}$ if $x>x_m$ and 0 otherwise.
As $\beta$ increases, the disturbance is more concentrated on boundaries. This heavy-tailed distribution captures extreme disturbances, giving insight into the worst case performance.

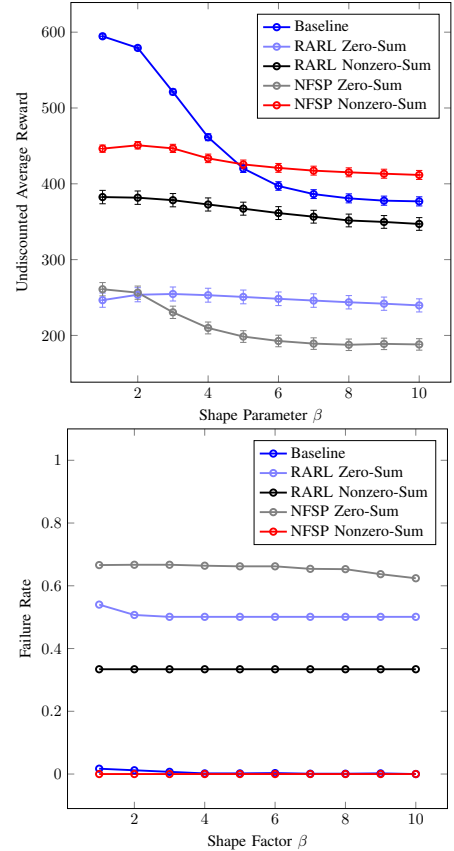
\begin{figure}[!t]
    \vspace{10pt}
    \centering
    \scalebox{.6}{\begin{tikzpicture}[]
\begin{axis}[height = {10cm}, legend pos = {north east}, ylabel = {Undiscounted Average Reward}, xlabel = {Shape Parameter $\beta$}, width = {10cm}]\addplot+ [
mark = {o}, blue,very thick, error bars/.cd, 
x dir=both, x explicit, y dir=both, y explicit]
table [
x error plus=ex+, x error minus=ex-, y error plus=ey+, y error minus=ey-
] {
x y ex+ ex- ey+ ey-
1.0 594.6064584461777 0.0 0.0 2.769303271352835 2.769303271352835
2.0 579.2208482869132 0.0 0.0 2.7841188417004576 2.7841188417004576
3.0 521.204241303304 0.0 0.0 3.6994043546077617 3.6994043546077617
4.0 461.4488748820357 0.0 0.0 4.567937705845685 4.567937705845685
5.0 420.2200975186125 0.0 0.0 5.234843541376472 5.234843541376472
6.0 397.10128604799684 0.0 0.0 5.645944013007438 5.645944013007438
7.0 386.41085099487566 0.0 0.0 5.838084702816659 5.838084702816659
8.0 380.80275097553243 0.0 0.0 5.94434727668069 5.94434727668069
9.0 377.7737909235718 0.0 0.0 5.997633696228596 5.997633696228596
10.0 376.8371818919727 0.0 0.0 6.00125551366802 6.00125551366802
};
\addlegendentry{Baseline}
\addplot+ [
mark = {o}, blue!50,very thick, error bars/.cd, 
x dir=both, x explicit, y dir=both, y explicit]
table [
x error plus=ex+, x error minus=ex-, y error plus=ey+, y error minus=ey-
] {
x y ex+ ex- ey+ ey-
1.0 246.5499228136119 0.0 0.0 9.403840891833143 9.403840891833143
2.0 253.6741932427571 0.0 0.0 9.300263815292121 9.300263815292121
3.0 254.65414919743105 0.0 0.0 9.240024741237233 9.240024741237233
4.0 253.07820553897926 0.0 0.0 9.166953704025623 9.166953704025623
5.0 250.79177943436858 0.0 0.0 9.073151804005319 9.073151804005319
6.0 248.3200227717056 0.0 0.0 8.976712565821652 8.976712565821652
7.0 246.0172735574302 0.0 0.0 8.883462531006538 8.883462531006538
8.0 243.75397783961668 0.0 0.0 8.79702654006564 8.79702654006564
9.0 241.8896734353593 0.0 0.0 8.731195899140276 8.731195899140276
10.0 239.62183876960466 0.0 0.0 8.6437159846612 8.6437159846612
};
\addlegendentry{RARL Zero-Sum}
\addplot+ [
mark = {o}, black,very thick, error bars/.cd, 
x dir=both, x explicit, y dir=both, y explicit]
table [
x error plus=ex+, x error minus=ex-, y error plus=ey+, y error minus=ey-
] {
x y ex+ ex- ey+ ey-
1.0 382.5018301610325 0.0 0.0 8.84000551812277 8.84000551812277
2.0 381.7069473391918 0.0 0.0 8.823323800512155 8.823323800512155
3.0 378.3976801443249 0.0 0.0 8.757401510322204 8.757401510322204
4.0 372.7320996783086 0.0 0.0 8.6529704717244 8.6529704717244
5.0 367.201207397784 0.0 0.0 8.571177244719577 8.571177244719577
6.0 361.3971922261446 0.0 0.0 8.499767298654696 8.499767298654696
7.0 356.6112184380539 0.0 0.0 8.453486911952913 8.453486911952913
8.0 351.6383530026259 0.0 0.0 8.416702243221287 8.416702243221287
9.0 349.70267180522353 0.0 0.0 8.406796096371153 8.406796096371153
10.0 347.03239324563117 0.0 0.0 8.393469545043398 8.393469545043398
};
\addlegendentry{RARL Nonzero-Sum}
\addplot+ [
mark = {o}, black!50,very thick, error bars/.cd, 
x dir=both, x explicit, y dir=both, y explicit]
table [
x error plus=ex+, x error minus=ex-, y error plus=ey+, y error minus=ey-
] {
x y ex+ ex- ey+ ey-
1.0 260.9300764305777 0.0 0.0 8.787365005949745 8.787365005949745
2.0 256.42640057794046 0.0 0.0 8.683477580080352 8.683477580080352
3.0 230.49710590761887 0.0 0.0 8.140451770370134 8.140451770370134
4.0 209.8642699108489 0.0 0.0 7.824277702556634 7.824277702556634
5.0 198.5252908850576 0.0 0.0 7.683578835983534 7.683578835983534
6.0 192.68163775369507 0.0 0.0 7.619502741456687 7.619502741456687
7.0 189.21956769060534 0.0 0.0 7.599076012724782 7.599076012724782
8.0 187.66592483934525 0.0 0.0 7.579532310367809 7.579532310367809
9.0 188.87295892635723 0.0 0.0 7.547676793528324 7.547676793528324
10.0 188.13324426528754 0.0 0.0 7.5507264224382835 7.5507264224382835
};
\addlegendentry{NFSP Zero-Sum}
\addplot+ [
mark = {o}, red,very thick, error bars/.cd, 
x dir=both, x explicit, y dir=both, y explicit]
table [
x error plus=ex+, x error minus=ex-, y error plus=ey+, y error minus=ey-
] {
x y ex+ ex- ey+ ey-
1.0 446.36811109320496 0.0 0.0 4.78020821156688 4.78020821156688
2.0 450.84397922895647 0.0 0.0 4.900859595577781 4.900859595577781
3.0 446.7344095261738 0.0 0.0 5.374464387965646 5.374464387965646
4.0 433.5663704187178 0.0 0.0 5.637621922850695 5.637621922850695
5.0 425.5859169787137 0.0 0.0 5.774512593222685 5.774512593222685
6.0 420.9921223741091 0.0 0.0 5.848177127941492 5.848177127941492
7.0 417.3685168085841 0.0 0.0 5.888251085291631 5.888251085291631
8.0 415.1674255038421 0.0 0.0 5.911313026101256 5.911313026101256
9.0 413.3143602413668 0.0 0.0 5.937609432182113 5.937609432182113
10.0 411.68898644022886 0.0 0.0 5.947587713834818 5.947587713834818
};
\addlegendentry{NFSP Nonzero-Sum}
\end{axis}

\end{tikzpicture}}
    \scalebox{.6}{\begin{tikzpicture}[]
\begin{axis}[height = {10cm}, legend pos = {north east}, ylabel = {Failure Rate}, ymax = {1.1}, xlabel = {Shape Factor $\beta$}, width = {10cm}]\addplot+ [mark = {o}, blue,very thick]coordinates {
(1.0, 0.017)
(2.0, 0.012)
(3.0, 0.007)
(4.0, 0.002)
(5.0, 0.002)
(6.0, 0.003)
(7.0, 0.001)
(8.0, 0.001)
(9.0, 0.002)
(10.0, 0.0)
};
\addlegendentry{Baseline}
\addplot+ [mark = {o}, blue!50,very thick]coordinates {
(1.0, 0.54)
(2.0, 0.507)
(3.0, 0.501)
(4.0, 0.501)
(5.0, 0.501)
(6.0, 0.501)
(7.0, 0.501)
(8.0, 0.501)
(9.0, 0.501)
(10.0, 0.501)
};
\addlegendentry{RARL Zero-Sum}
\addplot+ [mark = {o}, black,very thick]coordinates {
(1.0, 0.334)
(2.0, 0.334)
(3.0, 0.334)
(4.0, 0.334)
(5.0, 0.334)
(6.0, 0.334)
(7.0, 0.334)
(8.0, 0.334)
(9.0, 0.334)
(10.0, 0.334)
};
\addlegendentry{RARL Nonzero-Sum}
\addplot+ [mark = {o}, black!50,very thick]coordinates {
(1.0, 0.666)
(2.0, 0.667)
(3.0, 0.667)
(4.0, 0.664)
(5.0, 0.662)
(6.0, 0.662)
(7.0, 0.654)
(8.0, 0.653)
(9.0, 0.637)
(10.0, 0.624)
};
\addlegendentry{NFSP Zero-Sum}
\addplot+ [mark = {o}, red,very thick]coordinates {
(1.0, 0.0)
(2.0, 0.0)
(3.0, 0.0)
(4.0, 0.0)
(5.0, 0.0)
(6.0, 0.0)
(7.0, 0.0)
(8.0, 0.0)
(9.0, 0.0)
(10.0, 0.0)
};
\addlegendentry{NFSP Nonzero-Sum}
\end{axis}

\end{tikzpicture}}
    \caption{\small Average path reward (top) and failure rate (bottom) of different policies under Pareto disturbance with increasingly heavy-tailed shape parameter, $\beta$.}
    \label{fig:ParetoScatter}
\end{figure}

\Cref{fig:ParetoScatter} gives the results under the disturbance described above. The horizontal-axis is the shape parameter $\beta$ and the vertical-axes are the average reward and collision rate. As the disturbance concentrates more on the boundary ($\beta$ increases), the reward of baseline drops more rapidly than other policies, which shows the robustness of game theoretic methods. Policies trained with nonzero-sum rewards show better performance than ones trained with zero-sum, demonstrating the influence of the additional adversary rewards. 
The NFSP policy for the nonzero-sum game implementation demonstrates the most robust performance and highest reward when the variance is high.

Further, the NFSP nonzero-setting policy exhibits the safest performance as well; the resulting policy always results in no collisions. 
The baseline policy performs nearly as well, despite more rapidly degrading performance in terms of expected reward. 

\subsection{Adversarial Disturbance Test}
Inspired by RARL \cite{pinto2017robust}, we test the robustness of the policy with adversarial disturbances. We train an adversary to apply disturbance on acceleration and steering while holding the protagonist's policy constant (for NFSP, this is the best response policy of the protagonist). The training setup is the same as RARL. We then test the protagonist under the disturbance generated by this adversary.

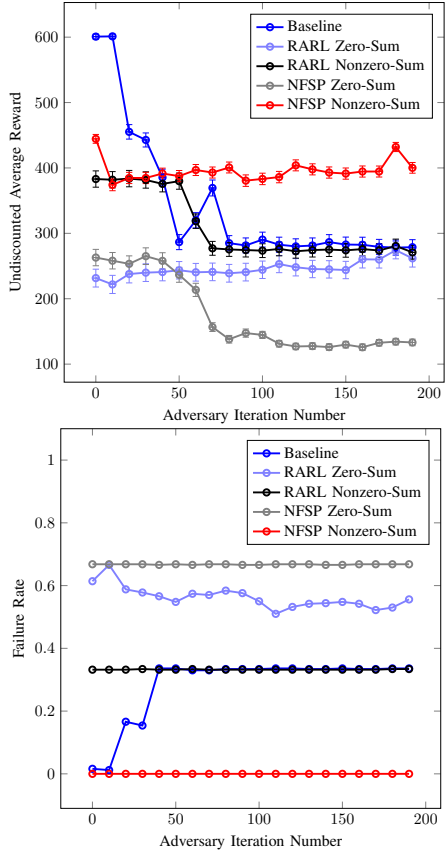
\begin{figure}[!t]
    \centering
    \vspace{10pt}
    \scalebox{.6}{\begin{tikzpicture}[]
\begin{axis}[height = {10cm}, legend pos = {north east}, ylabel = {Undiscounted Average Reward}, xlabel = {Adversary Iteration Number}, width = {10cm}]\addplot+ [
mark = {o}, blue,very thick, error bars/.cd, 
x dir=both, x explicit, y dir=both, y explicit]
table [
x error plus=ex+, x error minus=ex-, y error plus=ey+, y error minus=ey-
] {
x y ex+ ex- ey+ ey-
0.0 600.8042341428531 0.0 0.0 3.7564319489991402 3.7564319489991402
10.0 601.2115930994518 0.0 0.0 3.3620055117798215 3.3620055117798215
20.0 455.1809607381286 0.0 0.0 11.075264947031561 11.075264947031561
30.0 442.7955630277711 0.0 0.0 10.73334339218283 10.73334339218283
40.0 385.39837097688206 0.0 0.0 12.669949926876336 12.669949926876336
50.0 286.5949311901142 0.0 0.0 11.525045764264016 11.525045764264016
60.0 319.9060808687263 0.0 0.0 11.719793275155046 11.719793275155046
70.0 369.38339608401117 0.0 0.0 12.185710239324042 12.185710239324042
80.0 284.9563961647179 0.0 0.0 11.528720987147677 11.528720987147677
90.0 281.48825323106513 0.0 0.0 11.483392153670632 11.483392153670632
100.0 290.4350189640835 0.0 0.0 11.37756509989059 11.37756509989059
110.0 282.6006561127598 0.0 0.0 11.529148895298766 11.529148895298766
120.0 280.05651543528455 0.0 0.0 11.554176655352656 11.554176655352656
130.0 281.36199379719676 0.0 0.0 11.595835478756186 11.595835478756186
140.0 286.5331880349203 0.0 0.0 11.47041979357254 11.47041979357254
150.0 282.8650657715333 0.0 0.0 11.405992216547059 11.405992216547059
160.0 282.49892105384055 0.0 0.0 11.56616699501577 11.56616699501577
170.0 279.32199299651705 0.0 0.0 11.452951906804886 11.452951906804886
180.0 278.18635780139897 0.0 0.0 11.448232160111704 11.448232160111704
190.0 278.81285875068386 0.0 0.0 11.487969424924891 11.487969424924891
};
\addlegendentry{Baseline}
\addplot+ [
mark = {o}, blue!50,very thick, error bars/.cd, 
x dir=both, x explicit, y dir=both, y explicit]
table [
x error plus=ex+, x error minus=ex-, y error plus=ey+, y error minus=ey-
] {
x y ex+ ex- ey+ ey-
0.0 231.5515060482118 0.0 0.0 13.637935782161318 13.637935782161318
10.0 221.87800464914108 0.0 0.0 13.86037372990734 13.86037372990734
20.0 237.76654811605 0.0 0.0 13.56074773905397 13.56074773905397
30.0 239.9172609288454 0.0 0.0 13.506268560021265 13.506268560021265
40.0 240.819015027595 0.0 0.0 13.41729833845678 13.41729833845678
50.0 243.5498435854057 0.0 0.0 13.298866625089484 13.298866625089484
60.0 240.48181320869827 0.0 0.0 13.513272134807629 13.513272134807629
70.0 241.05315083821998 0.0 0.0 13.460639179892476 13.460639179892476
80.0 238.94846086962568 0.0 0.0 13.537638980335641 13.537638980335641
90.0 240.71388588051397 0.0 0.0 13.511528541687657 13.511528541687657
100.0 244.2891981161627 0.0 0.0 13.332901428365735 13.332901428365735
110.0 252.98239094819303 0.0 0.0 13.14599802796899 13.14599802796899
120.0 248.1327723327656 0.0 0.0 13.245905122465606 13.245905122465606
130.0 245.44713048418322 0.0 0.0 13.267612783186173 13.267612783186173
140.0 244.89526900864286 0.0 0.0 13.26761125563906 13.26761125563906
150.0 243.9081555737441 0.0 0.0 13.279009965895398 13.279009965895398
160.0 260.28687381439494 0.0 0.0 13.237640426009058 13.237640426009058
170.0 259.9566206194363 0.0 0.0 13.007689609496417 13.007689609496417
180.0 274.24674120433076 0.0 0.0 13.245972265124115 13.245972265124115
190.0 261.7632238844553 0.0 0.0 13.310081557803773 13.310081557803773
};
\addlegendentry{RARL Zero-Sum}
\addplot+ [
mark = {o}, black,very thick, error bars/.cd, 
x dir=both, x explicit, y dir=both, y explicit]
table [
x error plus=ex+, x error minus=ex-, y error plus=ey+, y error minus=ey-
] {
x y ex+ ex- ey+ ey-
0.0 383.03848149062554 0.0 0.0 12.465716280246161 12.465716280246161
10.0 381.95583971589474 0.0 0.0 12.447035167596248 12.447035167596248
20.0 383.6931011572385 0.0 0.0 12.514320821469495 12.514320821469495
30.0 381.598589797 0.0 0.0 12.493471759835856 12.493471759835856
40.0 375.63095439742625 0.0 0.0 12.23491458939196 12.23491458939196
50.0 379.811325837755 0.0 0.0 12.358482773651712 12.358482773651712
60.0 318.6043631875385 0.0 0.0 11.044860958363325 11.044860958363325
70.0 277.0876897187777 0.0 0.0 10.602112363926858 10.602112363926858
80.0 275.38909901612976 0.0 0.0 10.616298629341822 10.616298629341822
90.0 274.4277630262368 0.0 0.0 10.57792720674968 10.57792720674968
100.0 273.5165242292912 0.0 0.0 10.541850507501698 10.541850507501698
110.0 276.1292528651522 0.0 0.0 10.52260478889996 10.52260478889996
120.0 272.59207342711045 0.0 0.0 10.541400532606003 10.541400532606003
130.0 274.4200244223548 0.0 0.0 10.615230781314652 10.615230781314652
140.0 275.1958659061031 0.0 0.0 10.59324305630231 10.59324305630231
150.0 274.01213660966346 0.0 0.0 10.504148310560263 10.504148310560263
160.0 275.88465858790266 0.0 0.0 10.479404170346491 10.479404170346491
170.0 273.7639307226485 0.0 0.0 10.476349382650714 10.476349382650714
180.0 280.6721961539402 0.0 0.0 10.506936378771801 10.506936378771801
190.0 270.91109424407443 0.0 0.0 10.466761040387842 10.466761040387842
};
\addlegendentry{RARL Nonzero-Sum}
\addplot+ [
mark = {o}, black!50,very thick, error bars/.cd, 
x dir=both, x explicit, y dir=both, y explicit]
table [
x error plus=ex+, x error minus=ex-, y error plus=ey+, y error minus=ey-
] {
x y ex+ ex- ey+ ey-
0.0 262.9398426940747 0.0 0.0 12.535001419191154 12.535001419191154
10.0 258.1198988124198 0.0 0.0 12.55954359065696 12.55954359065696
20.0 253.3468784446368 0.0 0.0 12.19907570999506 12.19907570999506
30.0 265.0719718428392 0.0 0.0 12.746848003469973 12.746848003469973
40.0 257.91354943895954 0.0 0.0 12.426458176892032 12.426458176892032
50.0 236.58529374052722 0.0 0.0 11.417920742859094 11.417920742859094
60.0 213.55262808064128 0.0 0.0 10.075745360865554 10.075745360865554
70.0 156.63199166476582 0.0 0.0 6.683810521180209 6.683810521180209
80.0 138.059743679116 0.0 0.0 5.81305173194784 5.81305173194784
90.0 147.45958596640273 0.0 0.0 6.248491822982983 6.248491822982983
100.0 144.64932863809307 0.0 0.0 5.432807827803697 5.432807827803697
110.0 131.2353643865924 0.0 0.0 5.0283976973585744 5.0283976973585744
120.0 127.00608791876964 0.0 0.0 4.896187005926245 4.896187005926245
130.0 127.56319601187514 0.0 0.0 5.032179976781072 5.032179976781072
140.0 126.0683561810678 0.0 0.0 4.958221444009975 4.958221444009975
150.0 129.80892100912325 0.0 0.0 5.05718034335305 5.05718034335305
160.0 125.78837644411509 0.0 0.0 5.012758965727553 5.012758965727553
170.0 132.66314202465654 0.0 0.0 4.9919294780345185 4.9919294780345185
180.0 134.37065377939152 0.0 0.0 5.025886589212532 5.025886589212532
190.0 133.1627101852742 0.0 0.0 5.036218330853636 5.036218330853636
};
\addlegendentry{NFSP Zero-Sum}
\addplot+ [
mark = {o}, red,very thick, error bars/.cd, 
x dir=both, x explicit, y dir=both, y explicit]
table [
x error plus=ex+, x error minus=ex-, y error plus=ey+, y error minus=ey-
] {
x y ex+ ex- ey+ ey-
0.0 444.4724714856513 0.0 0.0 6.747340873679373 6.747340873679373
10.0 374.18538352215876 0.0 0.0 9.018064865283096 9.018064865283096
20.0 384.6931613264479 0.0 0.0 8.599908197063435 8.599908197063435
30.0 384.69309889741044 0.0 0.0 8.619002983185561 8.619002983185561
40.0 391.44758351678985 0.0 0.0 8.262512834826504 8.262512834826504
50.0 387.5084909866462 0.0 0.0 8.590104699059829 8.590104699059829
60.0 396.84104672769126 0.0 0.0 8.419666298306723 8.419666298306723
70.0 392.9857439712037 0.0 0.0 8.44626821775788 8.44626821775788
80.0 400.6850690196291 0.0 0.0 8.331892569525575 8.331892569525575
90.0 380.59016245204486 0.0 0.0 8.78292229262795 8.78292229262795
100.0 383.22080013194494 0.0 0.0 8.760838329196527 8.760838329196527
110.0 386.0130784077554 0.0 0.0 8.694930693945679 8.694930693945679
120.0 403.96325488781264 0.0 0.0 8.403675962117633 8.403675962117633
130.0 397.84836950196785 0.0 0.0 8.465684662654303 8.465684662654303
140.0 392.96682945848295 0.0 0.0 8.557564002053702 8.557564002053702
150.0 391.37545811507744 0.0 0.0 8.558411922858282 8.558411922858282
160.0 394.52770172817696 0.0 0.0 8.487778569508203 8.487778569508203
170.0 394.5268733098637 0.0 0.0 8.40811350013648 8.40811350013648
180.0 432.04283272698905 0.0 0.0 6.988822391484315 6.988822391484315
190.0 400.16066431366363 0.0 0.0 8.18254398772133 8.18254398772133
};
\addlegendentry{NFSP Nonzero-Sum}
\end{axis}

\end{tikzpicture}}
    \scalebox{.6}{\begin{tikzpicture}[]
\begin{axis}[height = {10cm}, legend pos = {north east}, ylabel = {Failure Rate}, ymax = {1.1}, xlabel = {Adversary Iteration Number}, width = {10cm}]\addplot+ [mark = {o}, blue,very thick]coordinates {
(0.0, 0.016)
(10.0, 0.012)
(20.0, 0.166)
(30.0, 0.154)
(40.0, 0.336)
(50.0, 0.336)
(60.0, 0.33)
(70.0, 0.33)
(80.0, 0.334)
(90.0, 0.334)
(100.0, 0.334)
(110.0, 0.336)
(120.0, 0.336)
(130.0, 0.334)
(140.0, 0.334)
(150.0, 0.336)
(160.0, 0.334)
(170.0, 0.334)
(180.0, 0.336)
(190.0, 0.336)
};
\addlegendentry{Baseline}
\addplot+ [mark = {o}, blue!50,very thick]coordinates {
(0.0, 0.614)
(10.0, 0.666)
(20.0, 0.588)
(30.0, 0.578)
(40.0, 0.566)
(50.0, 0.548)
(60.0, 0.574)
(70.0, 0.57)
(80.0, 0.584)
(90.0, 0.576)
(100.0, 0.55)
(110.0, 0.51)
(120.0, 0.532)
(130.0, 0.542)
(140.0, 0.544)
(150.0, 0.548)
(160.0, 0.542)
(170.0, 0.522)
(180.0, 0.53)
(190.0, 0.556)
};
\addlegendentry{RARL Zero-Sum}
\addplot+ [mark = {o}, black,very thick]coordinates {
(0.0, 0.332)
(10.0, 0.332)
(20.0, 0.332)
(30.0, 0.334)
(40.0, 0.332)
(50.0, 0.332)
(60.0, 0.334)
(70.0, 0.332)
(80.0, 0.332)
(90.0, 0.332)
(100.0, 0.332)
(110.0, 0.332)
(120.0, 0.332)
(130.0, 0.332)
(140.0, 0.332)
(150.0, 0.332)
(160.0, 0.332)
(170.0, 0.332)
(180.0, 0.334)
(190.0, 0.334)
};
\addlegendentry{RARL Nonzero-Sum}
\addplot+ [mark = {o}, black!50,very thick]coordinates {
(0.0, 0.668)
(10.0, 0.668)
(20.0, 0.668)
(30.0, 0.668)
(40.0, 0.666)
(50.0, 0.668)
(60.0, 0.666)
(70.0, 0.668)
(80.0, 0.668)
(90.0, 0.666)
(100.0, 0.666)
(110.0, 0.668)
(120.0, 0.668)
(130.0, 0.668)
(140.0, 0.666)
(150.0, 0.666)
(160.0, 0.668)
(170.0, 0.668)
(180.0, 0.668)
(190.0, 0.668)
};
\addlegendentry{NFSP Zero-Sum}
\addplot+ [mark = {o}, red,very thick]coordinates {
(0.0, 0.0)
(10.0, 0.0)
(20.0, 0.0)
(30.0, 0.0)
(40.0, 0.0)
(50.0, 0.0)
(60.0, 0.0)
(70.0, 0.0)
(80.0, 0.0)
(90.0, 0.0)
(100.0, 0.0)
(110.0, 0.0)
(120.0, 0.0)
(130.0, 0.0)
(140.0, 0.0)
(150.0, 0.0)
(160.0, 0.0)
(170.0, 0.0)
(180.0, 0.0)
(190.0, 0.0)
};
\addlegendentry{NFSP Nonzero-Sum}
\end{axis}

\end{tikzpicture}}
    \caption{\small Average path reward (top) and failure rate (bottom) of different policies under adversarial disturbance in which an adversary iteratively trains against the protagonist.}
    \label{fig:AdTest}
\end{figure}

\Cref{fig:AdTest} shows the average path rewards and collision rate as a function of the adversary's training iterations. 
The policy trained by NFSP with $r_a=3$ in the nonzero-sum setting shows strong robustness to the adversarial disturbance.
The failure rate remains zero, while the expected reward is approximately 400. 
The baseline policy initially has a high average reward and zero collision rate, but its performance drops rapidly as the adversary becomes stronger. 
Policies trained with RARL result in rewards similar to the baseline, while the policy with non-zero formulations exhibit better performance with respect to the failure rate. 
These results illustrate how reformulating adversarial learning as a nonzero-sum game improves overall performance. 

\section{Robustness to Model Mismatch}
\label{sec:realworld results}

We apply the policy to the real-world dynamics described in the previous section.  
Using the same metrics as before,  we quantify the trade-offs between the policies. 
We test the policies on a limited steer change model that approximates the real vehicle behavior (LSC test), and a setting where we change the axle distance for the vehicle (DA test).

\subsection{Limited Steer Change Test}
In this test, we limit the steer change of the ego vehicle in one time step (\SI{0.05}{\second}) to \ang{4.5}. We also add a uniformly distributed disturbance in the same limit as previous tests.
This test shows the policies robustness against model error. 

\begin{figure}[!t]
    \vspace{10pt}
    \centering
    \scalebox{.8}{\begin{tikzpicture}[]
\begin{axis}[ylabel = {Undiscounted Average Reward}, xmin = {-0.05}, xmax = {1.0}, ymax = {700.0}, xlabel = {Failure Rate}, ymin = {0.0}]\addplot+[scatter, scatter src=explicit symbolic, only marks = {true}, scatter/classes = {{a={mark=triangle,blue},b={mark=triangle,blue!50},c={mark=triangle,black},d={mark=triangle,black!50},e={mark=triangle,red}}}, very thick] coordinates {
(0.202, 464.7952774276263) [a]
(0.568, 238.90997456945067) [b]
(0.64, 168.22873678432714) [c]
(0.668, 264.6293603142665) [d]
(0.04, 434.93034159067724) [e]
};
\addlegendentry{Baseline}
\addlegendentry{RARL Zero-Sum}
\addlegendentry{RARL Nonzero-Sum}
\addlegendentry{NFSP Zero-Sum}
\addlegendentry{NFSP Nonzero-Sum}
\end{axis}

\end{tikzpicture}}
    \caption{\small Scatter plot showing the trade-off between average path reward and failure rate of different policies under limited steer change constraint and uniform disturbance. }
    \label{fig:LSCScatter}
\end{figure}
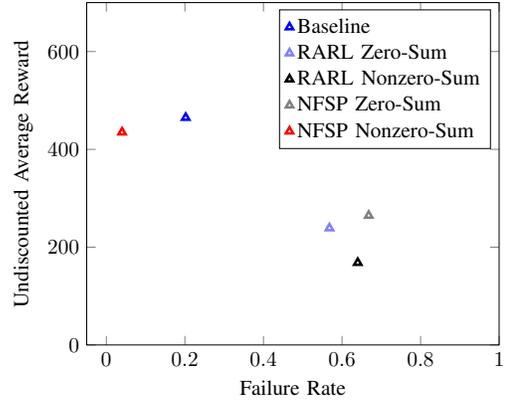

As shown in \cref{fig:LSCScatter}, the proposed methods shows a reduced average reward compared to the baseline.
While many of these adversarial learning methods show degraded safety, the NFSP solution in the nonzero-sum formulation improves safety with only a slightly degraded expected reward. 

\subsection{Different Axle Distance Test}
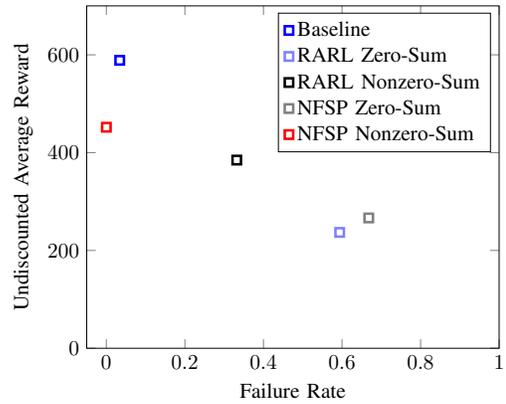
\begin{figure}[!t]
    \centering
    \scalebox{.8}{\begin{tikzpicture}[]
\begin{axis}[ylabel = {Undiscounted Average Reward}, xmin = {-0.05}, xmax = {1.0}, ymax = {700.0}, xlabel = {Failure Rate}, ymin = {0.0}]\addplot+[scatter, scatter src=explicit symbolic, only marks = {true}, scatter/classes = {{a={mark=square,blue},b={mark=square,blue!50},c={mark=square,black},d={mark=square,black!50},e={mark=square,red}}}, very thick] coordinates {
(0.034, 588.7434894060091) [a]
(0.594, 236.86462746056085) [b]
(0.332, 384.7720544657187) [c]
(0.668, 266.4633508990834) [d]
(0.0, 451.9286502894849) [e]
};
\addlegendentry{Baseline}
\addlegendentry{RARL Zero-Sum}
\addlegendentry{RARL Nonzero-Sum}
\addlegendentry{NFSP Zero-Sum}
\addlegendentry{NFSP Nonzero-Sum}
\end{axis}

\end{tikzpicture}}
    \caption{\small Scatter plot showing the trade-off between average path reward and failure rate of different policies under different axle distance and uniform disturbance.}
    \label{fig:DAScatter}
\end{figure}

In this test, we vary the vehicle's axle distance $l_a$ and $l_b$ uniformly from \num{0.5} to \SI{2.5}{\meter}. Recall that during training, the axle distance is fixed to \SI{1}{\meter}. Again, uniformly distributed disturbance is also added to the inputs. 
The results are shown in \cref{fig:DAScatter}.

This test exhibits similar results as was previously observed. 
While many of the adversarial methods do not improve performance relative to the baseline, the NFSP nonzero-sum formulation improves safety with only a slight sacrifice in expected reward.

\subsection{Discussion}
\label{sec:disc}
\begin{figure}[h!]
\vspace{7pt}
    \centering
    \scalebox{.8}{\begin{tikzpicture}[]
\begin{axis}[height = {10cm}, legend pos = {north east}, ylabel = {Undiscounted Average Reward}, xmin = {-0.05}, xmax = {1.0}, ymax = {700.0}, xlabel = {Failure Rate}, ymin = {0.0}, width = {10cm}]\addplot+ [no marks,blue,very thick]coordinates {
(-1, -1)
(-2, -2)
};
\addlegendentry{Baseline}
\addplot+ [no marks,blue!50,very thick]coordinates {
(-1, -1)
(-2, -2)
};
\addlegendentry{RARL Zero-Sum}
\addplot+ [no marks,black,very thick]coordinates {
(-1, -1)
(-2, -2)
};
\addlegendentry{RARL Nonzero-Sum}
\addplot+ [no marks,black!50,very thick]coordinates {
(-1, -1)
(-2, -2)
};
\addlegendentry{NFSP Zero-Sum}
\addplot+ [no marks,red,very thick]coordinates {
(-1, -1)
(-2, -2)
};
\addlegendentry{NFSP Nonzero-Sum}
\addplot+[scatter, scatter src=explicit symbolic, only marks = {true}, scatter/classes = {{a={mark=o,black},b={mark=square,black},c={mark=triangle,black}}}, very thick] coordinates {
(-1, -1) [a]
(-2, -2) [b]
(-3, -3) [c]
};
\addlegendentry{Pareto $\beta=10$}
\addlegendentry{DA}
\addlegendentry{LSC}
\addplot+[scatter, scatter src=explicit symbolic, only marks = {true}, scatter/classes = {{a={mark=o,blue},b={mark=o,blue!50},c={mark=o,black},d={mark=o,black!50},e={mark=o,red}}}, very thick] coordinates {
(0.0, 376.8371818919727) [a]
(0.501, 239.62183876960466) [b]
(0.334, 347.03239324563117) [c]
(0.624, 188.13324426528754) [d]
(0.0, 411.68898644022886) [e]
};
\addplot+[scatter, scatter src=explicit symbolic, only marks = {true}, scatter/classes = {{a={mark=square,blue},b={mark=square,blue!50},c={mark=square,black},d={mark=square,black!50},e={mark=square,red}}}, very thick] coordinates {
(0.034, 588.7434894060091) [a]
(0.594, 236.86462746056085) [b]
(0.332, 384.7720544657187) [c]
(0.668, 266.4633508990834) [d]
(0.0, 451.9286502894849) [e]
};
\addplot+[scatter, scatter src=explicit symbolic, only marks = {true}, scatter/classes = {{a={mark=triangle,blue},b={mark=triangle,blue!50},c={mark=triangle,black},d={mark=triangle,black!50},e={mark=triangle,red}}}, very thick] coordinates {
(0.202, 464.7952774276263) [a]
(0.568, 238.90997456945067) [b]
(0.64, 168.22873678432714) [c]
(0.668, 264.6293603142665) [d]
(0.04, 434.93034159067724) [e]
};
\end{axis}

\end{tikzpicture}}
    \caption{\small Scatter plot showing the trade-off between average path reward and failure rate of different policies for all tests.}
    \label{fig:TogetherScatter}
    \vspace{-7pt}
\end{figure}
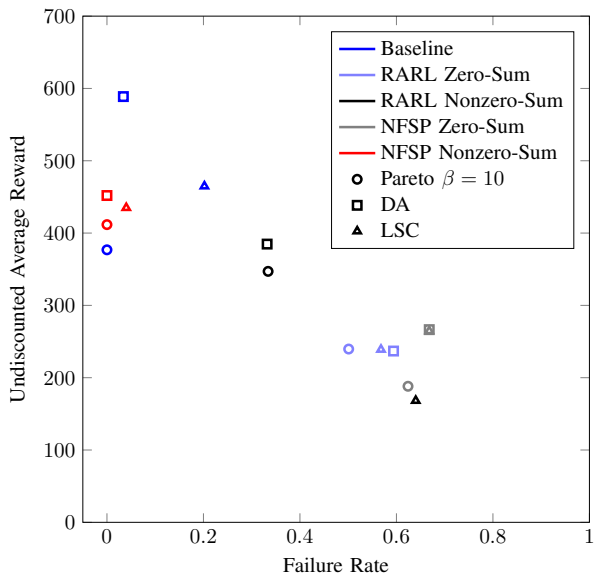

To summarize the findings of this work, \cref{fig:TogetherScatter} illustrates the efficiency and safety trade-off for tested methods across all validation tests. 
The color indicates the policy and formulation and the shape indicates the robustness test. 

Generally, the performance variance of policies trained with robust learning methods is smaller than that of the baseline.
We also observe that augmenting previous solution methods with a nonzero-sum formulation significantly improves robustness with respect to efficiency and safety.
Overall, the NFSP method in the nonzero-sum formulation results exhibit the best performance with regard to maintaining safety with minimal sacrifice in efficiency.

Interestingly, most robust learning methods degrade both efficiency and safety, exhibiting worst performance than baseline.
This phenomenon is likely a result of the game theoretic formulation and solution methods converging on suboptimal equilibria.
Thus, our hypothesis that using robust learning methods improves robustness and efficiency is confirmed, but only by using semi-competitive games and the NFSP solution method. 

\section{Conclusions}
\label{sec:conc}

In this paper, we tested a variety of the two player Markov game formulations for robust reinforcement learning, varying the training methods (RARL and NFSP) and reward structure (zero-sum and nonzero-sum with additional adversary reward). The resulting policies were validated under different disturbance distributions and different vehicle dynamic models. 
Under all tests, the NFSP with nonzero-sum rewards shows overall strong robustness, safety, and more conservative performance than the TRPO single agent learning baseline. This method is also significantly more consistent between tests than the other approaches, which implies that the policy is more generalizable.  

The other approaches (RARL variants and NFSP solving a zero-sum game) exhibited poor performance in both safety and efficiency metrics.  
This behavior is likely induced by the  adversary becoming too ``strong" for the current protagonist to recover from, and thus converges to a low performance point. 
However, with the addition of the cooperative reward and the averaging from fictitious self play, the protagonist and the adversary policies evolve more smoothly and converge more closely to the game's equilibrium. 
We will expand this work to a multi-player setting to capture a wider variety of disturbances and test more solution methods that can be applied to multi-player games like Policy-Space Response Oracles \cite{lanctot2017unified}. 
We will also explore the application of this robust training method on the test vehicle.

\section*{Acknowledgements}
We thank Alex Kuefler and Kunal Menda for inspiring this work and the members of the Stanford Intelligent Systems Laboratory for their help on the implementation. 


\bibliographystyle{IEEEtran}
\bibliography{references}

\clearpage

\end{document}